\documentclass[conference]{IEEEtran}
\IEEEoverridecommandlockouts
\usepackage{cite}
\usepackage{amsmath,amssymb,amsfonts}
\usepackage{multirow}
\usepackage{subfigure}
\usepackage{makecell}
\usepackage{algorithm}
\usepackage{algorithmic}
\usepackage{graphicx}
\usepackage{textcomp}
\usepackage{xcolor}
\usepackage{hyperref}
\usepackage{url}
\usepackage{color}
\usepackage{booktabs}

\newcommand{\norm}[1]{\left\lVert#1\right\rVert}
\newcommand{\R}{\mathbb{R}}
\newcommand*{\argmin}{\operatornamewithlimits{argmin}\limits}

\def\BibTeX{{\rm B\kern-.05em{\sc i\kern-.025em b}\kern-.08em
    T\kern-.1667em\lower.7ex\hbox{E}\kern-.125emX}}
\begin{document}

\title{Deep Stable Representation Learning on Electronic Health Records}

% \author{\IEEEauthorblockN{Yingtao Luo}
% \IEEEauthorblockA{
% \textit{Carnegie Mellon University}\\
% Pittsburgh, USA \\
% yingtaoluo@cmu.edu}
% \and
% \IEEEauthorblockN{Zhaocheng Liu}
% \IEEEauthorblockA{\textit{Kuaishou Technology} \\
% Beijing, China \\
% lio.h.zen@gmail.com}
% \and
% \IEEEauthorblockN{Qiang Liu}
% \IEEEauthorblockA{\textit{Chinese Academy of Sciences} \\
% Beijing, China \\
% qiang.liu@nlpr.ia.ac.cn}
% }

\author{
    \IEEEauthorblockN{Yingtao Luo\IEEEauthorrefmark{2}, Zhaocheng Liu\IEEEauthorrefmark{3}, Qiang Liu\IEEEauthorrefmark{4}\IEEEauthorrefmark{1}}
    \IEEEauthorblockA{\IEEEauthorrefmark{2}Carnegie Mellon University, Pittsburgh, USA
    \\yingtaoluo@cmu.edu}
    \IEEEauthorblockA{\IEEEauthorrefmark{3}Kuaishou Technology, Beijing, China
    \\lio.h.zen@gmail.com}
    \IEEEauthorblockA{\IEEEauthorrefmark{4}Center for Research on Intelligent Perception and Computing, National Laboratory of Pattern Recognition, \\Institute of Automation, Chinese Academy of Sciences, Beijing, China
    \\qiang.liu@nlpr.ia.ac.cn}
    
    \thanks{\IEEEauthorrefmark{1}To whom correspondence should be addressed.}
}

% \author{\IEEEauthorblockN{Anonymous}}

\maketitle

\begin{abstract}
Deep learning models have achieved promising disease prediction performance of the Electronic Health Records (EHR) of patients. However, most models developed under the I.I.D. hypothesis fail to consider the agnostic distribution shifts, diminishing the generalization ability of deep learning models to Out-Of-Distribution (OOD) data. In this setting, spurious statistical correlations that may change in different environments will be exploited, which can cause sub-optimal performances of deep learning models. The unstable correlation between procedures and diagnoses existed in the training distribution can cause spurious correlation between historical EHR and future diagnosis. To address this problem, we propose to use a causal representation learning method called Causal Healthcare Embedding (CHE). CHE aims at eliminating the spurious statistical relationship by removing the dependencies between diagnoses and procedures. We introduce the Hilbert-Schmidt Independence Criterion (HSIC) to measure the degree of independence between the embedded diagnosis and procedure features. Based on causal view analyses, we perform the sample weighting technique to get rid of such spurious relationship for the stable learning of EHR across different environments. Moreover, our proposed CHE method can be used as a flexible plug-and-play module that can enhance existing deep learning models on EHR. Extensive experiments on two public datasets and five state-of-the-art baselines unequivocally show that CHE can improve the prediction accuracy of deep learning models on out-of-distribution data by a large margin. In addition, the interpretability study shows that CHE could successfully leverage causal structures to reflect a more reasonable contribution of historical records for predictions.

\end{abstract}

\maketitle

\section{Introduction}
Healthcare predictive model for healthcare disease diagnosis based on Electronic Health Records (EHR) is a key engine for improving the quality of clinical care. \cite{charles2013adoption}. In the US, nearly 96\% of hospitals had a digital electronic health records (EHR) in 2015 \cite{charles2013adoption}, which emphasizes the importance of learning EHR. The comprehensive patient information (such as demographics, diagnoses, and procedures) in EHR provides valuable assistance for personal health status tracking and monitoring \cite{cheng2016risk, ma2017dipole, ma2018health, zhang2019metapred, gao2020stagenet}. To predict the future diagnoses based on a patient's historical EHR, many deep learning models \cite{choi2016retain, choi2017gram, gao2019camp, ma2020concare} are proposed with promising accuracy to discover the statistical correlations in the training distribution for predictions. 

Despite their great successes, the challenge of the out-of-distribution (OOD) problem has not yet been fully addressed in previous works, which may cause sub-optimal learning performance on EHR. The I.I.D. hypothesis that most models are built upon does not hold true for practical situations due to the inevitable distribution shifts such as data selection bias and confounding factors \cite{huang2006correcting, brookhart2010confounding, hendrycks2018benchmarking, bengio2019meta}. We present the spurious correlation between diagnoses and procedures as a representative example of out-of-distribution problems in electronic health records. The diagnoses and procedures are often correlated as clinicians select treatments according to the patients' current and historical diagnosis records based on medical experience and knowledge. However, the patients' demographics and insurance information may vary a lot in the training and test datasets, causing the subtle correlation between diagnoses and procedures to vary in different environments. We argue that the EHR prediction models may be misled by the subtle dependency between diagnoses and procedures, resulting in spurious correlations between historical EHR and future prediction that are unstable when confronting the OOD data in practice. As a result, the learned statistical correlations cannot guarantee to be as effective on inference as on the training dataset. 

\begin{figure}
\centering
\includegraphics[width=0.42\textwidth]{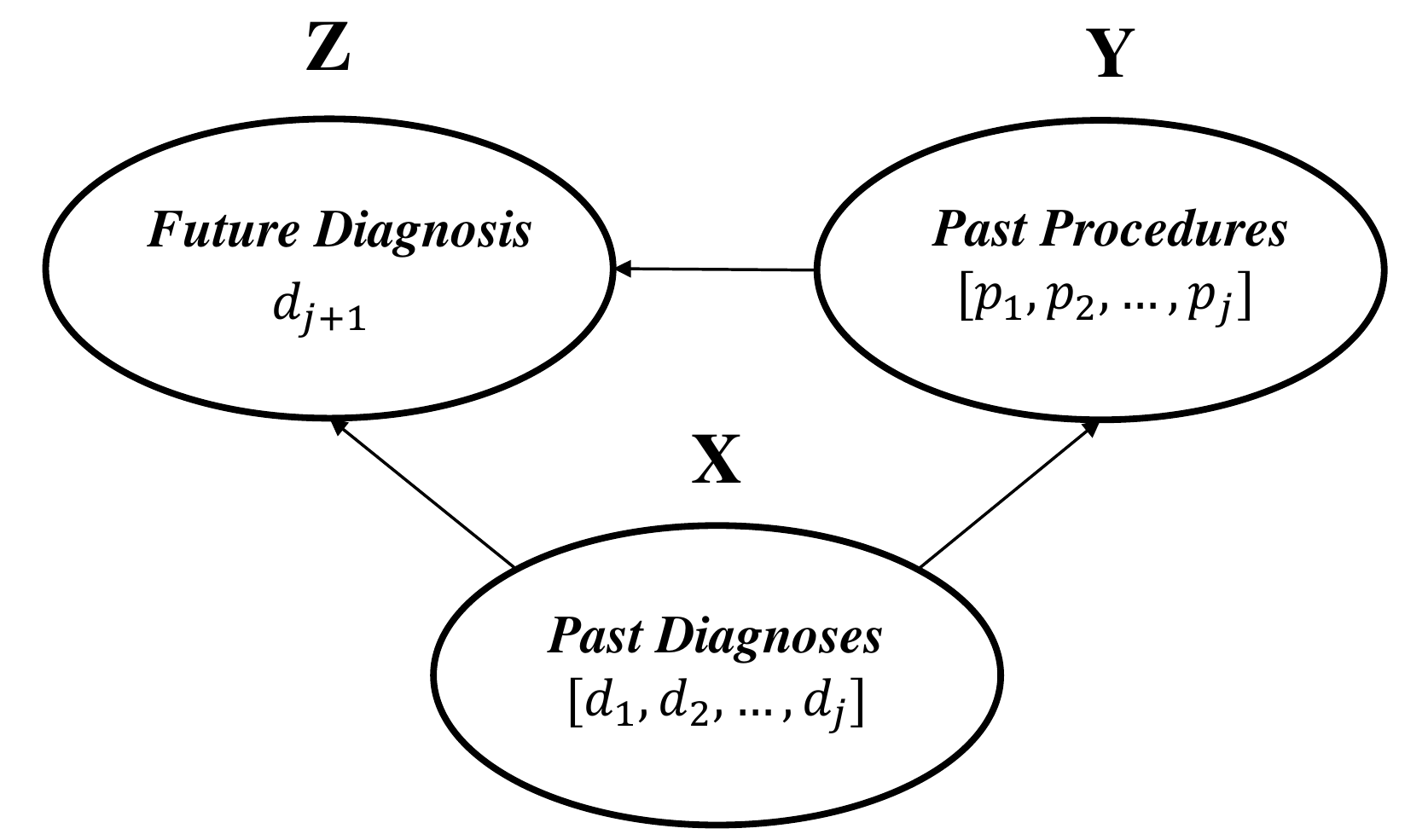}
\caption{The causal diagram of diagnosis prediction in EHR.}
\label{fig:diagram}
\end{figure}

In the following, we present the causal view analyses to discuss how the correlation between diagnoses and procedures as an example can cause the spurious relationship for model prediction. As shown in Fig. \ref{fig:diagram}, the causal diagram of future diagnosis prediction in EHR consists of two sequences of features, i.e. ``past diagnoses'' ($X$) and ``past procedures'' ($Y$). By the ignorability assumption in causality \cite{pearl2009causal}, any other potential confounders are considered uncorrelated to both diagnoses and procedures. Therefore, these potential confounders will not be discussed for a better understanding of the scheme.
Here, we discuss the case where the $X$ and $Y$ are correlated, thus the causal effect of each feature cannot be accurately estimated by deep learning models.
In EHR, each diagnosis has an impact on the current and future procedures.
At the same time, both diagnoses and procedures influence the ``future diagnosis'' ($Z$).
Because doctors give treatments based on the same medical knowledge, diagnoses and procedures are strongly correlated.
Due to the strong correlation of $X \to Y$, it is hard for deep learning models to learn a stable relationship of $X \to Z$ and/or $Y \to Z$.
As an instance, patients diagnosed with diabetes can take insulin, and diabetes may cause puffiness.
With the strong correlation between diabetes and insulin, a machine learning model has a great chance to learn that insulin causes puffiness.
Moreover, the correlation between the two variables may be different in various data distributions, which causes difficulty for model generalization to Out-Of-Distribution (OOD) data.
For example, procedures may vary among different insurance types and only some of them can cover insulin drugs, which may result in diet control treatment or non-insulin drug treatments such as Exenatide and Liraglutide for some patients with the same diagnoses.
% For example, unprofessional clinics might tend to rely solely on the current diagnosis to make a procedure while clinicians at professional city hospitals might pay more attention to the medical history as a whole to make decisions.
Therefore, models trained by one type of insurances may not always generalize to new insurances.

An intuitive approach to address this problem is to remove the dependency between $X$ and $Y$, so that this selection bias in the training dataset will not affect the inference phase. In our case, if the model does not exploit the correlation between diabetes and insulin in a selected training dataset for predicting puffiness, it will find that $X \rightarrow Z$ is the truly stable relationship that reflects causation. To this end, we are interested in a method that discovers stable correlations reflecting causal effect of each feature across different environments, which is free of data biases introduced by the distribution shifts between training and inference. Such a method demonstrating the ability to find a more causal model can point out that potential of future studies and deployments on various machine learning healthcare systems. Sadly, many algorithms such as domain generalization \cite{muandet2013domain}, causal transfer learning \cite{rojas2018invariant} and invariant causal prediction \cite{peters2016causal} cannot deal with distribution shifts unobserved in the training data. 

To obtain a stable correlation structure between each variable and the final prediction, a strand of variable decorrelation technique \cite{kuang2018stable, kuang2020stable} is proposed for linear models. Its basic notion is to remove the dependencies between variables through a sample weighting method and make the correlation structure between each variable and the prediction free of the confounding factors of other variables. In Fig. \ref{fig:diagram}, the arrow from each diagnosis $X$ to each procedure $Y$ will be removed, which leaves the causal diagram with independent $X$ and $Y$ to accurately estimate their contributions. While the concept of variable decorrelation is tempting for healthcare systems, how to extend it to a high-dimensional nonlinear deep learning model with sequential data can be difficult. There are two challenges to tackle. First, with nonlinear neural layers, the nonlinear correlation in deep learning healthcare cannot be measured and eliminated by linear methods. Second, the sample weighting should be redefined to accommodate the sequential data that any past diagnoses can have an impact on a future procedure along the time. It is vital to efficiently remove the dependencies of all combinations of diagnoses and procedures without excessive computational complexity.

In this paper, we propose a causal representation learning method for sequential diagnosis prediction in EHR, called Causal Healthcare Embedding (\textbf{CHE}). To address the two challenges, first, we use Hilbert Schmidt Independence Criterion (HSIC) \cite{gretton2007kernel, greenfeld2020robust} that measures the norm of cross-covariance from $X$ to $Y$, the degree of dependence between $X$ and $Y$ for feature decorrelation \cite{bahng2020learning}, which can align with the nonlinear neural models. By minimizing HSIC($X$,$Y$), i.e. the degree of dependence between diagnoses and procedures, we expect that the model will get rid of the spurious relationships that are hard to generalize to OOD data. Second, as pointed out by \cite{zou2020counterfactual}, treatments can be represented by latent factors as an alternative for estimating causality. While it is computationally expensive and inaccurate to calculate the binary sample weighting for all sequential combinations of treatments\cite{arbour2021permutation}, we apply HSIC on the two low-dimensional latent representations generated by diagnoses and procedures. By minimizing the HSIC in the loss function, the dependency between $X$ and $Y$ can be minimized throughout the training. Without spurious relationships caused by unstable correlations, deep learning models can exploit the causation between each feature and the prediction. Moreover, the learning of causation can improve the model generalization to different environments without necessarily knowing the training data a priori. We perform the proposed CHE on multiple baseline models for predicting the future diagnosis and show by extensive experiments that it can improve these models by large margins.

To be noted, our proposed CHE method is a plug-and-play module for the diagnosis prediction task.
It can be easily and adaptively incorporated with various diagnosis prediction models.
To summarize, our contributions are listed as follows:
\begin{itemize}
\item For the first time, we propose a causal representation learning method for healthcare diagnosis prediction, which removes the dependencies between variables such as diagnoses and procedures by sample weighting on the latent representation.
\item We show that the proposed method can learn a stable correlation between each causal feature to the prediction, which makes predictions stable across different data distributions without knowing the training data a priori.
\item Our proposed method is a plug-and-play module that can work well with diagnosis prediction models in a variety of scenarios. We prove that its computational complexity is identical to model training, without the concerns of estimating a exponentially growing number of treatments caused by the combinatorial nature of sequential data.
\item Extensive experiments on public datasets show that our method increases both NDCG and ACC of five state-of-the-art baselines by a significant margin, especially when applying to data of different sources and/or distributions. 
\end{itemize}

\section{Related Work}
In this section, we review the existing works for mining the EHR data, especially the state-of-the-art models on disease diagnosis prediction.
Moreover, we introduce some related works on counterfactual prediction and variable decorrelation.

\subsection{EHR Data Mining}
The mining of EHR is essential for improving the healthcare management of patients. Many tasks that aim at improving healthcare quality can be identified as EHR data mining, such as risk prediction \cite{cheng2016risk, che2017boosting, ma2018health, zhang2019metapred, luo2022learning}, disease progression \cite{choi2015constructing, zhang2019attain, ye2020lsan}, phenotyping \cite{che2015deep, liu2015temporal, zeng2018natural}, diagnosis prediction \cite{ma2018kame, gao2019camp, choi2020learning}.
Owing to the sequential pattern of EHR data, Recurrent Neural Networks (RNNs) are naturally suitable, and Long Short-Term Memory (LSTM) \cite{hochreiter1997long} has been successfully applied.
RETAIN \cite{choi2016retain} presents a reverse time attention model that preserves interpretability.
Dipole \cite{ma2017dipole} incorporates bidirectional RNN for making a prediction based on EHR.
Camp \cite{gao2019camp} uses demography information in co-attention model for diagnosis prediction.
ConCare \cite{ma2020concare} proposes to incorporate multi-head self-attention to model the sequential data of EHR.
StageNet \cite{gao2020stagenet} integrates time intervals between visits into LSTM to model the stages of health conditions.
INPREM \cite{zhang2020inprem} applies Bayesian neural network in an attention-based prediction model for improving the model interpretability.
HiTANet \cite{luo2020hitanet} proposes hierarchical time-aware attention networks for health risk prediction.
LSAN \cite{ye2020lsan} combines both long- and short-term information in EHR to make predictions.
SETOR \cite{peng2021sequential} utilizes ontological representation and neural ordinary equation for diagnosis prediction.
Meanwhile, multi-sourced data is also considered in recent works \cite{zhang2021learning,chen2021unite}.
Besides, constructed on sequential prediction models, medical knowledge graphs are modeled to provide some prior knowledge for more accurate predictions \cite{choi2017gram,ma2018kame,zhang2020hierarchical,ye2021medpath}.

\subsection{Counterfactual Prediction}

Counterfactual learning \cite{johansson2016learning} is an important direction of research in causal inference \cite{pearl2009causal, morgan2015counterfactuals}. Counterfactual learning can enable people to estimate the probability of counterfactual events and eventually identify the unbiased causal relationships between events. The existing counterfactual learning approaches usually reweight samples based on propensity scores \cite{rosenbaum1983central, austin2011introduction, hassanpour2019learning, lopez2017estimation}, which indicate the probabilities of observation under different environments.
Under the binary treatment setting, balancing the sample weights in the loss function can remove confounding bias to make causal prediction \cite{hassanpour2019counterfactual, arbour2021permutation}. Recent works further extend counterfactual learning to the multi-level treatment \cite{yoon2018ganite} and bundle treatment \cite{zou2020counterfactual} settings. Meanwhile, Permutation Weighting (PW) \cite{arbour2021permutation} conducts permutation on observed features for calculating propensity scores. These methods directly control input variables, therefore, the accurate estimation of propensity score when the number of treatments in sequential data is growing exponentially can be a challenge.

\subsection{Variable Decorrelation}

Recently, stable learning methods have been proposed to perform variable decorrelation for learning causal features in models from biased data.
Stable learning can be viewed as another perspective of causal learning technique, in which there is no implicit treatments and the distribution of unobserved samples is unknown \cite{kuang2018stable}.
Existing stable learning methods are mostly investigated in linear models.
Specifically, most methods conduct stable learning via decorrelation among features of samples, which tries to  make the feature distribution closer to independently identically distribution \cite{kuang2018stable,kuang2020stable,shen2020stable0}.
Sample Reweighted Decorrelation Operator (SRDO) \cite{shen2020stable} generates some unobserved samples, and trains a binary classifier to get the probabilities of observation for weighting the observed samples.
Recently, some works investigate to conduct stable learning on neural networks, such as on convolutional neural networks \cite{zhang2021deep} and graph neural networks \cite{fan2021generalizing}.

\section{Preliminary}

In this section, we first formulate the diagnosis prediction problem. Then, we discuss causality in EHR data.

\subsection{Problem Formulation}

In the EHR data, we have a set of patients $V = \{v_1, v_2, ..., v_{|V|}\}$, and patient $v_i$ has $t^i$ visits.
Diagnoses and procedures are both represented in International Classification of Diseases, Ninth Revision (ICD-9)\footnote{\url{https://www.cdc.gov/nchs/icd/icd9.htm}} medical codes, where we have $M$ unique diagnosis medical codes and $N$ unique procedure medical codes.
For each patient $v_i$ with $j$ visits, there exists a historical diagnosis sequence $D^i_j = \left[d^i_1, d^i_2, ..., d^i_j\right]$ and a historical procedure sequence $P^i_j = \left[p^i_1, p^i_2, ..., p^i_j\right]$.
Each diagnosis and procedure are $M$-dimensional multi-hot vector and $N$-dimensional multi-hot vector respectively, which means that $d^i_j \in \{0,1\}^M$ and $p^i_j \in \{0,1\}^N$, where $1 \le j \le t^i$.
In this work, we would like to predict future diagnoses, i.e., predicting what diseases a patient will have in the future, based on historical EHR.
Specifically, in this work, given $D^i_j$ and $P^i_j$, we need to predict future diagnosis $d^i_{j+1}$. The descriptions of notations are shown in Table \ref{notation}.

\begin{table}
\centering
\caption{Notations and Descriptions}
\begin{tabular}{c c c}
\toprule
  \textbf{Notation} & \textbf{Description} \\
  \midrule
  $V$ & The set of patients in the EHR data\\
  $v_i$ & The $i$-th patient in the EHR data\\
  $t^i$ & The number of visits of the $i$-th patient \\
  $D^i_j$ & The diagnoses of the $i$-th patient until the $j$-th visit \\
  $d^i_j$ & The diagnosis of the $i$-th patient at the $j$-th visit \\
  $P^i_j$ & The procedures of the $i$-th patient until the $j$-th visit\\
  $p^i_j$ & The procedure of the $i$-th patient at the $j$-th visit\\
  $\textbf{E}^{i,j}_D$ & The embedding of the historical diagnoses $D^i_j$\\
  $\textbf{E}^{i,j}_d$ & The embedding of the diagnosis $d^i_j$ \\
  $\textbf{E}^{i,j}_P$ & The embedding of the historical procedures $D^i_j$\\
  $\textbf{E}^{i,j}_p$ & The embedding of the procedure $p^i_j$ \\
  $\textbf{W}$ & The sample weights for all patients\\
  $\omega^i_j$ & The sample weight of the $i$-th patient at the $j$-th visit\\
  $r$ & The dimensionality of the embedding space\\
\bottomrule
\label{notation}
\end{tabular}
\end{table}

\begin{figure}
\centering
\includegraphics[width=0.45\textwidth]{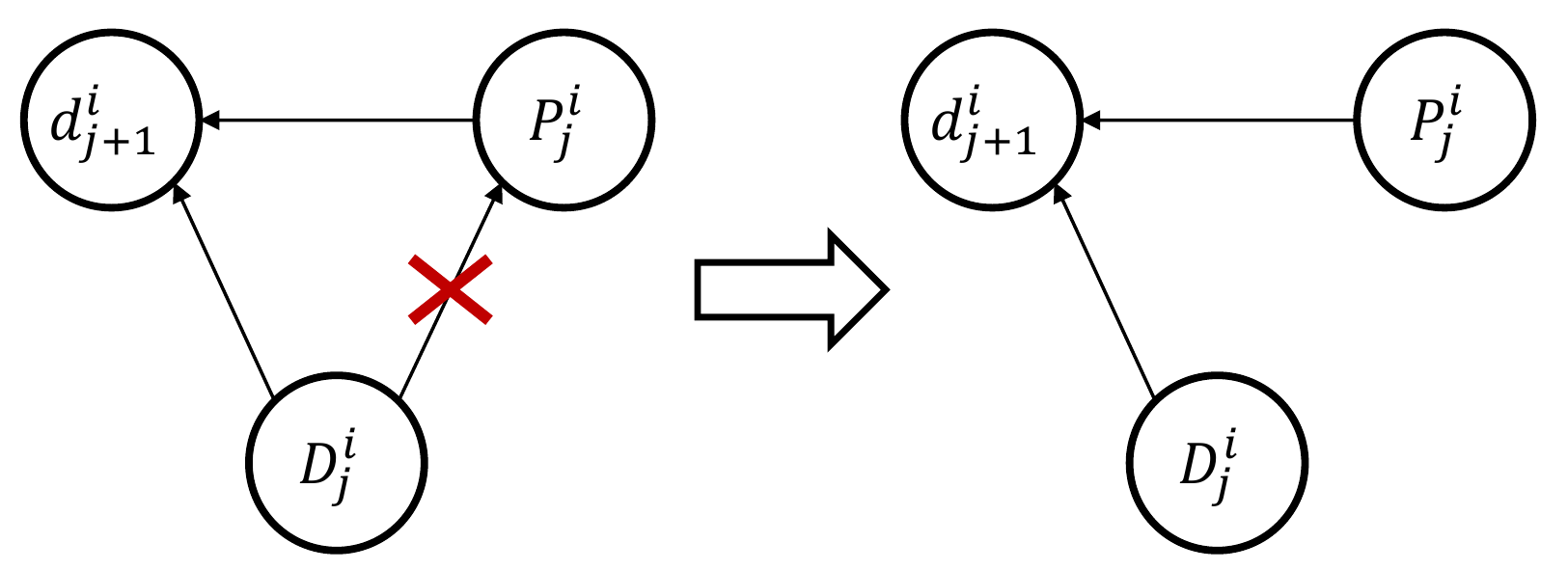}
\caption{Decorrelation between past diagnosis sequence $D^i_j$ and past procedure sequence $P^i_j$ until the $j$-th visit for patient $v_i$, for the more accurate and stable prediction of the diagnosis $d^i_{j+1}$ of a future time $j+1$. }
\label{fig:decorrelation}
\end{figure}

\begin{figure*}
\centering
\includegraphics[width=0.9\textwidth]{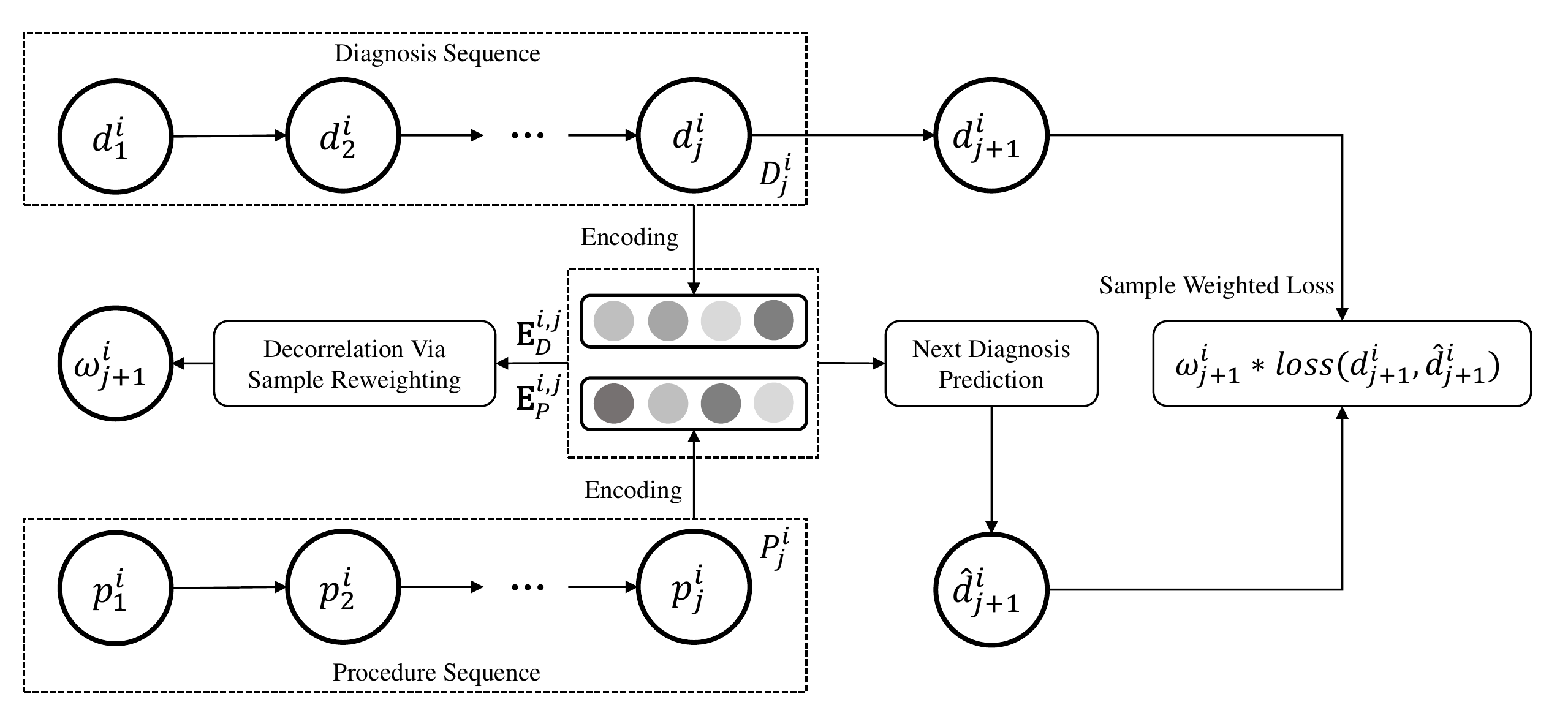}
\caption{The schematic diagram of decorrelation between past diagnoses $D^i_j$ and past procedures $P^i_j$ via sample weighting. }
\label{fig:model}
\end{figure*}

\subsection{Causal View Analysis}
Several causal links in medical knowledge affect the causal relationships in EHR. 
First, historical diagnoses can affect future procedures, in which doctors select procedures according to the patients' historical data based on medical experience.
Second, both historical diagnoses and procedures can have impacts on future diagnoses, since 1) diseases have development processes that are accompanied by complications and 2) procedures affect the development and rehabilitation of diseases.
Consequently, these relationships lead to a strong correlation between the two sequences of historical diagnoses $D^i_j$ and procedures $P^i_j$, which we denote as $D^i_j \rightarrow P^i_j$.
Both diagnoses and procedures can have causal effects on future diagnosis, i.e. $D^i_j \to d^i_{j+1}$ and $P^i_j \to d^i_{j+1}$, as shown in Fig. 1. Causal inference theory \cite{pearl2009causal, morgan2015counterfactuals} requires that the conditional covariance between "Treatment" and "Covariates" to be zero. In our case, both diagnoses and procedures can be treatment(s) and/or covariates, so we ensure the conditional covariance of $D^i_j \rightarrow P^i_j$ to be zero.

The strong correlation between diagnoses and procedures hinders machine learning models from learning causal relationships for future diagnosis.
Without the ability to remove $D^i_j \rightarrow P^i_j$, machine learning models may only rely on $D^i_j$ or $P^i_j$ to predict $d^i_{j+1}$ depending on the training data distribution.
For example, patients with diabetes usually take insulin to cure the disease.
Diabetes may cause some complications, such as retinopathy and cataract, even when insulin is taken.
If diabetes and insulin frequently occur together, it is hard for deep learning models to identify whether diabetes or insulin causes complications. 
If the model weighs on the occurrence of insulin for predicting complications, it cannot generalize to scenarios when diabetes and insulin do not occur together that often. On the other hand, if the correlation between $D^i_j$ and $P^i_j$ is eliminated, the model can estimate the importance of diabetes and insulin for future diagnoses independently, as shown in Fig. 2. As changing insulin to other prescriptions neither influence the distribution of diabetes (the correlation is eliminated) nor the distribution of complications (the correlation does not exist in medicine), the model will rely solely on diabetes to predict possible complications. 

\subsection{The Effect of Correlation in Embedding Space}
The correlation is easy to understand in linear models, considering that multicollinearity problem has been widely studied in linear models \cite{farrar1967multicollinearity}.
Diagnosis prediction models are mostly based on deep learning approaches, thus the discussion should be done in the embedding space.
We first encode diagnoses and procedures into an embedding space by any EHR encoders
\begin{equation}
\mathop {\bf{E}}\nolimits_D^{i,j}  = Encoder\left( {\mathop D\nolimits_j^i } \right),
\end{equation}
\begin{equation}
\mathop {\bf{E}}\nolimits_P^{i,j}  = Encoder\left( {\mathop P\nolimits_j^i } \right),
\end{equation}

For diagnosis prediction, we can learn a deep learning model $f\left( \cdot \right)$ that satisfies
\begin{equation} \label{eq:deep}
\mathop {d}\nolimits^i_{j+1}  = f\left( \mathop {\bf{E}}\nolimits_D^{i,j} , \mathop {\bf{E}}\nolimits_P^{i,j} \right).
\end{equation}
Considering the strong correlation between $D^i_j$ and $P^i_j$, we can predict one of their embeddings based on another as
\begin{equation}
\mathop {\bf{E}}\nolimits_D^{i,j}  = \mathop g\nolimits_{P \to D} \left( \mathop {\bf{E}}\nolimits_P^{i,j} \right) + \mathop \varepsilon \nolimits_{P \to D},
\end{equation}
\begin{equation}
\mathop {\bf{E}}\nolimits_P^{i,j}  = \mathop g\nolimits_{D \to P} \left( \mathop {\bf{E}}\nolimits_D^{i,j} \right) + \mathop \varepsilon \nolimits_{D \to P},
\end{equation}
where $\mathop g\nolimits_{P \to D} \left(  \cdot  \right)$ and $\mathop g\nolimits_{D \to P} \left(  \cdot  \right)$ are deep learning models, $\mathop \varepsilon \nolimits_{P \to D}$ and $\mathop \varepsilon \nolimits_{D \to P}$ are residual errors. $\mathop g\nolimits_{P \to D} \left(  \cdot  \right)$ and $\mathop g\nolimits_{D \to P} \left(  \cdot  \right)$ are both deep learning models, so their parameters can be merged with $f\left( \cdot \right)$.
Therefore, Eq. (\ref{eq:deep}) can rewritten as
\begin{equation}
\mathop {d}\nolimits^i_{j+1}  = f\left( {\mathop \varepsilon \nolimits_{P \to D}, \mathop {\mathop {\bf{E}}\nolimits_P^{i,j}} } \right),
\end{equation}
or
\begin{equation}
\mathop {d}\nolimits^i_{j+1} = f\left( {\mathop {\bf{E}}\nolimits_D^{i,j} ,\mathop \varepsilon \nolimits_{D \to P} } \right).
\end{equation}

Due to the strong correlation between $D^i_j$ and $P^i_j$, the model can rely on either $\mathop {\bf{E}}\nolimits_D^{i,j}$ or $\mathop {\bf{E}}\nolimits_P^{i,j}$ to make diagnosis prediction, which is unstable when the correlation changes.
Thus, we need to decorrelate past diagnoses and past procedures to obtain $\mathop g\nolimits_{P \to D} \left( {\mathop {\bf{E}}\nolimits_P^{i,j}} \right) = 0$ and $\mathop g\nolimits_{D \to P} \left( {\mathop {\bf{E}}\nolimits_D^{i,j}} \right) = 0$.
Accordingly, as illustrated in Fig. \ref{fig:decorrelation}, we plan to remove $\mathop {\bf{E}}\nolimits_D^{i,j} \rightarrow \mathop {\bf{E}}\nolimits_P^{i,j}$ to learn more accurate, stable and causal diagnosis prediction models.
Then, the contribution of $D^i_j$ and $P^i_j$ to predicting $d^i_{j+1}$ can be free of the interference of unstable $D^i_j \rightarrow P^i_j$.

\section{Methodology} \label{sec:method}
In this section, we introduce the sample weighting method with independence testing statistics to conduct causal disease diagnosis prediction, which is a plug-and-play module that aligns well with deep learning models.

\subsection{Hilbert Schmidt Independence Criterion}
The removal of dependencies between $\mathop {\bf{E}}\nolimits_D^{i,j}$ and $\mathop {\bf{E}}\nolimits_P^{i,j}$ is at the core of sample weighting. To measure the dependency for optimization, we introduce HSIC \cite{gretton2007kernel, greenfeld2020robust}, an independence testing statistics as the Hilbert-Schmidt norm of the cross-covariance operator between the distributions in Reproducing Kernel Hilbert Space (RKHS). Consider the measurable, positive definite kernel $k$ of variables and the corresponding RKHS $H$ \cite{sriperumbudur2009kernel}. For all $h_D \in H_D, h_P \in H_P$, the cross-covariance operator $\Sigma_{DP}$ from $H_D$ to $H_P$ is:
\begin{align}
\langle h_D, \Sigma_{DP} h_P \rangle = \mathbb{E}_{DP} [h_D(\mathop {\bf{E}}\nolimits_D^{i,j}) h_P(\mathop {\bf{E}}\nolimits_P^{i,j})] \nonumber \\
- \mathbb{E}_{D} [h_D(\mathop {\bf{E}}\nolimits_D^{i,j})] \mathbb{E}_{P} [h_P(\mathop {\bf{E}}\nolimits_P^{i,j})].
\end{align}
As proved by \cite{zhang2021deep}, if the product of $k_D$ and $k_P$ is characteristic, $\mathbb{E}[k_D(\mathop {\bf{E}}\nolimits_D^{i,j}, \mathop {\bf{E}}\nolimits_D^{i,j})] < \infty$ and $\mathbb{E}[k_P(\mathop {\bf{E}}\nolimits_P^{i,j}, \mathop {\bf{E}}\nolimits_P^{i,j})] < \infty$, we have
\begin{align}
\Sigma_{DP} = 0 \Longleftrightarrow D \perp P,
\end{align}
which means that if $\textbf{E}^{i,j}_D$ cannot be transformed into $\textbf{E}^{i,j}_P$ via a nonlinear operator, the two variables are independent. 

The squared Hilbert-Schmidt norm of the cross-covariance operator $\Sigma_{DP}$ can be approximated by the unbiased calculation in the embedding space as
\begin{align}
\text{HSIC}(\textbf{E}_D,\textbf{E}_P) = \frac{1}{|V|(t^i-1)} \sum_{i=1}^{|V|} \sum_{j=1}^{t^i-1} \text{HSIC}_{local}(\textbf{E}^{i,j}_D,\textbf{E}^{i,j}_P).
\end{align}
% \begin{align}
% HSIC(\textbf{D},\textbf{P}) = \frac{1}{(n-1)^2} Tr(K_DJK_PJ),
% \end{align}
Specifically, if $r$ denotes the hidden dimensionality, we can calculate the HSIC of $\textbf{E}^{i,j}_D \in \R^{r}$ and $\textbf{E}^{i,j}_P \in \R^{r}$ by
\begin{align}
\text{HSIC}_{local}(\textbf{E}^{i,j}_D,\textbf{E}^{i,j}_P) = \frac{1}{(r-1)^2} Tr(K_dJK_pJ),
\end{align}
where $Tr$ is the trace of a matrix, $J=I-1/r$ with $I$ as an r-order identity matrix. $K_D$ and $K_P$ are any kernel matrices. We can consider RBF kernel to calculate
\begin{align}
K_d(x_1, x_2) = \exp(-\frac{\norm{x_1-x_2}_2^2}{\sigma^2}),
\end{align}
where $x_1, x_2 \in \textbf{E}^{i,j}_D$ represent the values in different dimensions of the latent representation. Therefore, $x_q \in \R^1$ for $\forall q \in [1, ..., r]$. Similarly, there is $K_p(x_1, x_2)$ where $x_1, x_2 \in \textbf{E}^{i,j}_P$ represent different dimensions of the latent representation. The kernel tricks of $K_d, K_p \in \R^{r \times r}$ can approximately calculate HSIC rapidly. In this way, for each time's visit by each patient, the cross-covariance in the embedding space from diagnoses to procedures can be measured by HSIC.

\subsection{Loss Functions}
Inspired by feature decorrelation techniques \cite{shen2020stable, shen2020stable0}, we propose to minimize HSIC by the sample weighting technique to mitigate the dependency between diagnoses and procedures in the embedded space. We use $\textbf{W}$ to denote sample weights, where the weight for patient $i$ at the $j$-th visit is denoted as $\omega^i_j$. We denote the weighted samples as $\textbf{WE}_D$ and $\textbf{WE}_P$, where the weighted samples of patient $i$ at the $j$-th visit are denoted as $\omega^i_j \textbf{E}^{i,j}_D$ and $\omega^i_j \textbf{E}^{i,j}_P$.

To minimize the correlation between diagnoses and procedures, we propose to optimize $\textbf{W}$ with HSIC as follows
\begin{align}
\textbf{W}^* = \argmin_{\textbf{W}} \text{HSIC}(\textbf{WE}_D,\textbf{WE}_P).
\end{align}
Meanwhile, we define the cross-entropy loss for the classification task of the diagnosis prediction as
\begin{align}
Enc, Prd &= \argmin_{Enc, Prd} \sum_{i=1}^{|V|} \sum_{j=1}^{t^i-1} \omega^i_j (n) \textbf{L}^i_j,
\end{align}
where
\begin{align} \label{eq:loss_pred}
\textbf{L}^i_j &= L( Prd(Enc(D^i_j),Enc(P^i_j)), d^{i}_{j+1}),
\end{align}
where $L$ denotes the cross-entropy loss function. $Enc$ represents the encoder that maps diagnoses and procedures into the embedding space. $Prd$ represents the final prediction layer that maps the latent representation into the one-hot probability vector. The architectures of $Enc$ and $Prd$ depend on the base model our method is used upon. 
$L$ is also based on the specific loss function used in the base model.

\subsection{Model Optimization}
We iteratively optimize the weighted loss and the HSIC by
\begin{align}
Enc_{n+1}, Prd_{n+1} &= \argmin_{Enc, Prd} \sum_{i=1}^{|V|} \sum_{j=1}^{t^i-1} \omega^i_j (n) \textbf{L}^i_j,
\end{align}
and
\begin{align}
\textbf{W}(n+1) &= \argmin_{\textbf{W}} \epsilon \cdot \text{HSIC}(\text{Diag},\text{Proc}),
\end{align}
where
\begin{align}
\text{Diag} = \textbf{W}(n) Enc_{n+1}(D^i_j), \text{Proc} = \textbf{W}(n) Enc_{n+1}(P^i_j).
\end{align}
$Enc_n$, $Prd_n$ and $\textbf{W}(n)$ indicates encoder, final prediction layer and sample weights at the $n$-th iteration, and $\textbf{W}{(0)}$ is initially set as ones. $\epsilon$ is a coefficient that balances the learning rates for updating the neural network and sample weights. 

Eq. (16) and Eq. (17) are optimized iteratively, meaning that we first optimize the neural network and then optimize the HSIC for each iteration. Every two subsequences $\textbf{E}^{i,j}_D$ and $\textbf{E}^{i,j}_P$ of length $j$ are fed into a neural network to calculate the cross-entropy loss, and sample weights are multiplied to the loss to update the model parameters. Then, we use the updated model to calculate the HSIC of $\textbf{E}^{i,j}_D$ and $\textbf{E}^{i,j}_P$ obtained by the encoder part of the model. The sample weighting reassigns the importance of each sample when calculating the loss function to remove the dependency between features.

To make the overall method presented in Section \ref{sec:method} clearer, we show the pseudo-code of training the CHE method in Algorithm \ref{alg_training}.

\begin{algorithm}
    \caption{The training process of CHE.}
    \label{alg_training}
    \begin{algorithmic}[1]
        \REQUIRE Set of patients $V = \{v_1, v_2, ..., v_{|V|}\}$, each $v_i$'s diagnosis sequence $D^i_{j^i} = \left[d^i_1, d^i_2, ..., d^i_{j^i}\right]$ and procedure sequence $P^i_{j^i} = \left[p^i_1, p^i_2, ..., p^i_{j^i}\right]$, maximum epoch $\mathop{Epoch}$, and any BaseModel. 
        \ENSURE Model parameters $Enc()$ and $Prd()$ in the BaseModel.
        \STATE Initialize epoch indicator $n \gets 0$.
        \STATE Initialize best epoch indicator $n_{best} \gets 0$.
        \STATE Initialize $Enc_0()$ and $Prd_0()$ randomly.
        \STATE Initialize sample weights $\textbf{W}$.
        For every $\omega_j^i \in \textbf{W}$, $\omega_j^i(0) \gets 1$ for $1 \le i \le |V|$ and $1 \le j \le t^i-1$.
        \WHILE{early-stopping not reached and $n < \mathop{Epoch}$.}
            \STATE Update model parameters $Enc_{n+1}()$ and $Prd_{n+1}()$ according to Eq. (18), while keeping $\textbf{W}(n)$ fixed.
            \STATE Update sample weights $\textbf{W}(n+1)$ according to Eq. (20), while keeping $Enc_{n+1}()$ and $Prd_{n+1}()$ fixed.
            \STATE $n \gets n+1$.
            \STATE Update $n_{best} \gets n$, if better result is achieved on the validation set.
        \ENDWHILE
        \RETURN $Enc_{n_{best}}()$ and $Prd_{n_{best}}()$.
    \end{algorithmic}
\end{algorithm}

% The convergence of both HSIC and weighted cross-entropy loss ensures that the fine-tuned model maps diagnoses and procedures into an embedding space where each diagnosis is independent of each procedure. 

\subsection{Complexity Analysis}
The time complexity of calculating HSIC only grows with the hidden dimensionality $r$. By naive algorithms, the multiplication of $K_d$ and $K_p$ is $O(r^3)$, and the calculation of trace is also $O(r^3)$. For deep learning models, $r$ is a hyperparameter and is thus trivial. The calculation of overall $\text{HSIC}$ is $O(|V|t)$ if we denote $t = max(t^i)$ for $\forall i$, which is linearly proportional to the number of visits in the data. This is acceptably efficient. On the other hand, the number of treatments for each timestamp is $M$, the number of unique ICD-9 codes for diagnosis. Considering the combination of ICD-9 codes in a sequence, the total number of treatments can be as many as $M^t$, which makes the traditional counterfactual weighting to estimate propensity scores very expensive. 

\section{Experiments}

In this section, we conduct extensive experiments to verify the effectiveness of our proposed CHE method.

\begin{table*}[t]
\centering
  \caption{With random data division, performances of BaseModels, PW+BaseModels and CHE+BaseModels. Best performances are indicated by bold fonts. The improvement indicates the relative increase of CHE+BaseModel over BaseModel. $*$ denotes significant improvement of CHE+BaseModel, measured by t-test with p-value$<0.01$, over BaseModel and PW+BaseModel.}
  \label{tab:main1}
\begin{tabular}{ccccccccccccc}
\toprule
\multirow{2}{*}{Approach} & \multicolumn{4}{c}{MIMIC-III} & \multicolumn{4}{c}{MIMIC-IV} & \multirow{2}{*}{Average} \\
                           & NDCG@10         & NDCG@20          & ACC@10         & ACC@20  
                           & NDCG@10         & NDCG@20          & ACC@10         & ACC@20 & \\
\midrule
LSTM       & 0.2648       & 0.2712        & 0.1779       & 0.2597
           & 0.3469       & 0.3386        & 0.2167       & 0.3084 & 0.2730\\
PW+LSTM         & 0.2669       & 0.2724        & 0.1791       & 0.2605        
                & 0.3488       & 0.3394        & 0.2177       & 0.3040 & 0.2736\\
CHE+LSTM         & $\ $\textbf{0.2756}$^*$       & $\ $\textbf{0.2809}$^*$        & $\ $\textbf{0.1853}$^*$       & $\ $\textbf{0.2690}$^*$       
                 & $\ $\textbf{0.3589}$^*$       & $\ $\textbf{0.3496}$^*$        & $\ $\textbf{0.2246}$^*$       & $\ $\textbf{0.3186}$^*$ & $\ $\textbf{0.2828}$^*$ \\
Improv     & 4.079\%      & 3.577\%        & 4.160\%       & 3.581\%
              & 3.459\%      & 3.249\%        & 3.646\%       & 3.307\% & 3.590\%\\
\midrule
RETAIN       & 0.3409       & 0.3413        & 0.2305       & 0.3261
             & 0.4095       & 0.3946        & 0.2568       & 0.3533 & 0.3316\\
PW+RETAIN         & 0.3436       & 0.3449        & 0.2316       & 0.3241
                  & 0.4120       & 0.3981        & 0.2580       & 0.3527 & 0.3331\\
CHE+RETAIN        & $\ $\textbf{0.3545}$^*$       & $\ $\textbf{0.3579}$^*$        & $\ $\textbf{0.2353}$^*$       & $\ $\textbf{0.3354}$^*$
                  & $\ $\textbf{0.4231}$^*$       & $\ $\textbf{0.4085}$^*$        & $\ $\textbf{0.2630}$^*$       & $\ $\textbf{0.3614}$^*$ & $\ $\textbf{0.3424}$^*$\\
Improv     & 3.989\%      & 4.864\%        & 2.082\%       & 2.852\%
              & 3.321\%      & 3.523\%        & 2.414\%       & 2.293\% & 3.257\%\\
\midrule
Dipole       & 0.3071       & 0.3104        & 0.2075       & 0.2959
             & 0.3801       & 0.3710        & 0.2379       & 0.3352 & 0.3056\\
PW+Dipole         & 0.3072       & 0.3110        & 0.2077       & 0.2965
                  & 0.3860       & 0.3754        & 0.2388       & 0.3376 & 0.3075\\
CHE+Dipole         & $\ $\textbf{0.3371}$^*$       & $\ $\textbf{0.3385}$^*$        & $\ $\textbf{0.2213}$^*$       & $\ $\textbf{0.3149}$^*$
                   & $\ $\textbf{0.4054}$^*$       & $\ $\textbf{0.3932}$^*$        & $\ $\textbf{0.2523}$^*$       & $\ $\textbf{0.3529}$^*$ & $\ $\textbf{0.3270}$^*$\\
Improv     & 9.769\%      & 8.842\%        & 6.651\%       & 6.421\%
              & 6.656\%      & 5.984\%        & 6.053\%       & 5.280\% & 7.003\%\\
\midrule
Concare       & 0.2963       & 0.2979        & 0.1949       & 0.2793
              & 0.3748       & 0.3615        & 0.2346       & 0.3226 & 0.2952\\
PW+Concare         & 0.2972       & 0.2980        & 0.1952       & 0.2798 
                   & 0.3720       & 0.3602        & 0.2335       & 0.3234 & 0.2948 \\
CHE+Concare         & $\ $\textbf{0.3068}$^*$       & $\ $\textbf{0.3121}$^*$        & $\ $\textbf{0.2076}$^*$       & $\ $\textbf{0.2935}$^*$
                    & $\ $\textbf{0.3876}$^*$       & $\ $\textbf{0.3760}$^*$        & $\ $\textbf{0.2444}$^*$       & $\ $\textbf{0.3371}$^*$ & $\ $\textbf{0.3081}$^*$\\
Improv     & 3.544\%      & 4.767\%        & 6.516\%       & 5.084\%
              & 3.415\%      & 4.011\%        & 4.177\%       & 4.495\% & 4.370\%\\
\midrule
Stagenet       & 0.3364       & 0.3379        & 0.2284       & 0.3222
               & 0.3979       & 0.3853        & 0.2513       & 0.3471 & 0.3258\\
PW+Stagenet         & 0.3343       & 0.3362        & 0.2267       & 0.3210
                    & 0.3960       & 0.3861        & 0.2529       & 0.3476 & 0.3251\\
CHE+Staegnet         & $\ $\textbf{0.3432}$^*$       & $\ $\textbf{0.3467}$^*$        & $\ $\textbf{0.2315}$^*$       & $\ $\textbf{0.3295}$^*$
                     & $\ $\textbf{0.4064}$^*$       & $\ $\textbf{0.3976}$^*$        & $\ $\textbf{0.2559}$^*$       & $\ $\textbf{0.3541}$^*$ & $\ $\textbf{0.3331}$^*$\\
Improv     & 2.021\%      & 2.604\%        & 1.357\%       & 2.266\% 
              & 2.136\%      & 3.192\%        & 1.830\%       & 2.017\% & 2.241\%\\
\bottomrule
\end{tabular}
\end{table*}

\subsection{Datasets}
We evaluate our proposed sequential stable learning method on two real-world datasets: MIMIC-III and MIMIC-IV. 
\begin{itemize}
\item \textbf{MIMIC-III Dataset} We use diagnoses and procedures data from the Medical Information Mart for Intensive Care (MIMIC-III) database\footnote{\url{https://physionet.org/content/mimiciii/1.4/}} \cite{johnson2016mimic}, which contains patients who stayed in critical care units of the Beth Israel Deaconess Medical Center between 2001 and 2012. Patients who had less than three admission records are excluded. After this preprocessing, the average number of visits for the 1970 selected patients is 3.69, the average number of codes in a visit is 13.23, the total number of unique ICD-9 codes in diagnoses is 3320, and the total number of unique ICD-9 codes in procedures is 988.

\item \textbf{MIMIC-IV Dataset} We use diagnoses and procedures data from the MIMIC-IV database\footnote{\url{https://physionet.org/content/mimiciv/0.4/}} \cite{johnson2016mimic}, which contains patients admitted between 2008 and 2019. Patients with less than three admission records are excluded. After this preprocessing, the average number of visits for the 10023 selected patients is 4.64, the average number of codes in a visit is 14.12, the total number of unique ICD-9 codes in diagnoses is 6274, and the total number of unique ICD-9 codes in procedures is 1973.
% \item \textbf{CMS Dataset} We use diagnoses and procedures from the Centers for Medicare and Medicaid Services (CMS) 2008-2010  medicare claims database\footnote{\url{https://www.cms.gov/Research-Statistics-Data-and-Systems/Downloadable-Public-Use-Files/SynPUFs/DE_Syn_PUF}}. Patients who had less than five admission records are excluded. After this preprocessing, the average number of visits for the 8534 selected patients is 5.69, the average number of codes in a visit is 8.50, the total number of unique ICD-9 codes in diagnoses is 4773, and the total number of unique ICD-9 codes in procedures is 3990.
\end{itemize}

\subsection{Baseline Models}
We apply our method on the following baselines for the overall evaluation of diagnosis prediction accuracy. For a fair comparison, all models are used with adaptation to our task where only historical diagnoses and procedures are available. Side information like ontology and temporal intervals is not fused, thus the performances may not necessarily match the ones reported in the original papers.
\begin{itemize}
\item \textbf{LSTM:} \cite{hochreiter1997long} A recurrent neural network with long-short term gating mechanism.
\item \textbf{RETAIN:} \cite{choi2016retain} A two-level neural model based on reverse time attention for healthcare.
\item \textbf{Dipole:} \cite{ma2017dipole} An attention-based bidirectional recurrent neural network for healthcare.
\item \textbf{Concare:} \cite{ma2020concare} A self-attention model with cross-head decorrelation to capture health context for healthcare.
\item \textbf{StageNet:} \cite{gao2020stagenet} A deep learning model with stage-aware LSTM and convolutional modules for health prediction.
\end{itemize}
We denote above models as \textbf{BaseModels}, and we incorporate them with the \textbf{CHE} method as \textbf{CHE+BaseModels}. We implement our method with the Mindspore framework.

In addition, counterfactual learning methods such as inverse propensity weighting are also frequently used. Therefore, we also conduct experiment on the counterfactual Permutation Weighting (\textbf{PW}) technique \cite{arbour2021permutation} for comparison. PW conducts permutation on observed features for calculating propensity scores. While the historical EHR in the dataset is regarded as positive samples, we randomly generate negative samples that do not exist in the dataset and estimate their propensity scores. The combination of various ICD-9 codes in a sequence is a large space as discussed in Section 3.2. We generate 10x larger number of negative samples to make the propensity estimation as accurate as possible.
We also incorporate PW with the above BaseModels, and name them as \textbf{PW+BaseModels}.
For each PW+BaseModel, the propensity scores are calculated via training the corresponding BaseModel with permutation.

\begin{table*}[t]
\centering
  \caption{Under out-of-distribution data, performances of BaseModels, PW+BaseModels and CHE+BaseModels. Best performances are indicated by bold fonts. The improvement indicates the relative increase of CHE+BaseModel over BaseModel. $*$ denotes significant improvement of CHE+BaseModel, measured by t-test with p-value$<0.01$, over BaseModel and PW+BaseModel.}
  \label{tab:main2}
\begin{tabular}{cccccccccccc}
\toprule
\multirow{2}{*}{Approach} & \multicolumn{4}{c}{MIMIC-III} & \multicolumn{4}{c}{MIMIC-IV} & \multirow{2}{*}{Average} \\
                        & NDCG@10         & NDCG@20          & ACC@10         & ACC@20
                        & NDCG@10         & NDCG@20          & ACC@10         & ACC@20\\
\midrule
LSTM        & 0.2082       & 0.2112        & 0.1420       & 0.1971
            & 0.4395       & 0.4149        & 0.2519       & 0.3571 & 0.2771\\
PW+LSTM         & 0.2075       & 0.2102        & 0.1413       & 0.2012
                & 0.4468       & 0.4221        & 0.2525       & 0.3598 & 0.2802\\
CHE+LSTM         & $\ $\textbf{0.2182}$^*$       & $\ $\textbf{0.2204}$^*$        & $\ $\textbf{0.1492}$^*$       & $\ $\textbf{0.2064}$^*$
                & $\ $\textbf{0.4649}$^*$       & $\ $\textbf{0.4390}$^*$        & $\ $\textbf{0.2692}$^*$       & $\ $\textbf{0.3792}$^*$ & $\ $\textbf{0.2933}$^*$\\
Improv     & 4.948\%      & 4.356\%        & 5.070\%       & 4.718\%
              & 5.779\%      & 5.809\%        & 6.868\%       & 6.189\% & 5.846\%\\
\midrule
RETAIN       & 0.2385       & 0.2447        & 0.1615      & 0.2364
             & 0.5195       & 0.4859        & 0.3019      & 0.4173 & 0.3257\\
PW+RETAIN         & 0.2396       & 0.2443        & 0.1614       & 0.2362
                  & 0.5251       & 0.4888        & 0.3042       & 0.4195 & 0.3274\\
CHE+RETAIN        & $\ $\textbf{0.2503}$^*$       & $\ $\textbf{0.2589}$^*$        & $\ $\textbf{0.1687}$^*$       & $\ $\textbf{0.2492}$^*$
                  & $\ $\textbf{0.5625}$^*$       & $\ $\textbf{0.5260}$^*$        & $\ $\textbf{0.3244}$^*$       & $\ $\textbf{0.4519}$^*$ & $\ $\textbf{0.3583}$^*$\\
Improv     & 4.948\%      & 5.803\%        & 4.458\%       & 5.415\%
              & 8.277\%      & 8.253\%        & 7.453\%       & 8.291\% & 6.939\%\\
\midrule
Dipole       & 0.2287       & 0.2367        & 0.1482       & 0.2264
             & 0.4741       & 0.4469        & 0.2770       & 0.3873 & 0.3032\\
PW+Dipole         & 0.2402       & 0.2481        & 0.1567       & 0.2354
                  & 0.4727       & 0.4433        & 0.2796       & 0.3865 & 0.3078\\
CHE+Dipole         & $\ $\textbf{0.2729}$^*$       & $\ $\textbf{0.2772}$^*$        & $\ $\textbf{0.1782}$^*$       & $\ $\textbf{0.2631}$^*$
                  & $\ $\textbf{0.5176}$^*$       & $\ $\textbf{0.4905}$^*$        & $\ $\textbf{0.3067}$^*$       & $\ $\textbf{0.4280}$^*$ & $\ $\textbf{0.3418}$^*$\\
Improv     & 19.33\%      & 17.11\%        & 20.24\%       & 16.21\%
              & 9.175\%      & 9.756\%        & 10.72\%       & 10.51\% & 12.73\%\\
\midrule
Concare       & 0.2139       & 0.2229        & 0.1398       & 0.2139
              & 0.4910       & 0.4616        & 0.2857       & 0.3974 & 0.3033\\
PW+Concare         & 0.2122       & 0.2216        & 0.1414       & 0.2146
                   & 0.4947       & 0.4628        & 0.2878       & 0.4002 & 0.3044\\
CHE+Concare         & $\ $\textbf{0.2253}$^*$       & $\ $\textbf{0.2382}$^*$        & $\ $\textbf{0.1521}$^*$       & $\ $\textbf{0.2299}$^*$
            & $\ $\textbf{0.5270}$^*$       & $\ $\textbf{0.4944}$^*$        & $\ $\textbf{0.3087}$^*$       & $\ $\textbf{0.4267}$^*$ & $\ $\textbf{0.3253}$^*$\\
Improv     & 5.330\%      & 6.864\%        & 8.798\%       & 7.480\%
              & 7.332\%      & 7.106\%        & 8.050\%       & 7.373\% & 7.254\%\\
\midrule
Stagenet       & 0.2149       & 0.2224        & 0.1456       & 0.2171
               & 0.5722       & 0.5423        & 0.3418       & 0.4754 & 0.3415\\
PW+Stagenet         & 0.2145       & 0.2230        & 0.1451       & 0.2153
                    & 0.5830       & 0.5522        & 0.3499       & 0.4874 & 0.3463\\
CHE+Staegnet         & $\ $\textbf{0.2295}$^*$       & $\ $\textbf{0.2373}$^*$        & $\ $\textbf{0.1544}$^*$       & $\ $\textbf{0.2320}$^*$
                    & $\ $\textbf{0.6861}$^*$       & $\ $\textbf{0.6567}$^*$        & $\ $\textbf{0.4269}$^*$       & $\ $\textbf{0.5879}$^*$ & $\ $\textbf{0.4014}$^*$\\
Improv     & 6.794\%      & 6.700\%        & 6.044\%       & 6.863\%
              & 19.91\%      & 21.10\%        & 24.90\%       & 23.66\% & 17.54\%\\
\bottomrule
\end{tabular}
\end{table*}

\subsection{Settings}
We conduct two diagnosis prediction experiments. In the first experiment, we aim at evaluating the performance of our proposed method when training data and test data are divided randomly by patients to approximately simulate I.I.D. distributions. Following prior works \cite{choi2016retain, ma2017dipole}, We randomly divide the dataset into the training, validation and testing set in a 0.75:0.1:0.15 ratio. In the second experiment, considering the insurance type, such as Medicare, Medicaid and Private, may affect procedures for similar diagnoses, we evaluate the performance when training and test data are divided by the type of insurances to simulate the scenario of OOD generalization. Here, we divide all the Medicare data into the training and validation set in a 0.7:0.3 ratio and use the Private/Other (MIMIC-III/MIMIC-IV) data as the test set.

Common hyperparameters used by all models in the experiments include learning rate, batch size, hidden dimension, dropout rate. CHE+BaseModel has a special hyperparameter: weighting coefficient $\epsilon$. These hyperparameters are tuned with an appropriate range to obtain the optimal evaluation metrics on the validation set for each individual model. The range of learning rate is \{1e-2, 3e-3, 1e-3\}, the range of batch size is \{16, 32, 64, 128, 256\}, the range of hidden dimension is \{16, 32, 64\}, the range of dropout rate is \{0,1, 0.5\}, the range of coefficient is \{0.1, 0.3, 1, 3, 10\}. We apply early-stopping so the training will stop if the validation metrics do not increase in twenty epochs and the test performance will be recorded. All results are averaged under five different random seeds and recorded in four significant figures. 
% Therefore, the rounding error is within $5 \times 10^{-5}$.

\subsection{Evaluation Metrics}
We adopt the top$k$ accuracy and normalized discounted cumulative gain (NDCG) to evaluate the diagnosis prediction performance. We use the same accuracy@$k$ metric used in prior works \cite{ma2017dipole, choi2017gram, zhang2020inprem}, which is defined as the correct medical ICD-9 codes ranked in top$k$ divided by min($k$, $|y_t|$), where $|y_t|$ is the number of ICD-9 codes in the ($t$+1)-th visit. NDCG@$k$ further considers the normalization of gains and the ranking of correct medical codes, where codes with higher relevance will affect the final score more than those with lower relevance.  
In our experiments, we use $k \in [10, 20]$. We also provide an averaged metric of all the four metrics on two datasets to reflect the overall comparison to baselines.

\subsection{Performance Comparison}
In this subsection, we conduct performance comparison among BaseModels, PW+BaseModels and CHE+BaseModels, from two perspectives: datasets with random data division and out-of-distribution division respectively.

First, Table \ref{tab:main1} shows performance comparison with random data division.
The results show that CHE can provide improvements under all metrics on both datasets, as CHE is designed to remove spurious relationships for models to focus on causal features.
Overall, the Averaged Metric of NDCG@$k$ and ACC@$k$ on two datasets increases by 4.092\% compared to BaseModels. We use t-test with a p-value of 0.01 to evaluate the performance improvement and confirm that the improvements by CHE on all BaseModels are statistically significant. 
Moreover, PW+BaseModels slightly outperform BaseModels, but CHE+BaseModels can still significantly outperform PW+BaseModels.
Using the optimization described in Eqs.(18-19), the HSIC is reduced by more than 1000 times. Because the dependency between past diagnoses and past procedures is minimized, non-causal features will not interfere with the model to learn from causal features for prediction. 

Second, in Table \ref{tab:main2}, we further show performance comparison with out-of-distribution data.
Because we use Medicare insurance data for training and Private/Other insurance data for testing, the training and test sets are not I.I.D. Therefore, this poses additional challenges to the generalization of deep learning models. The results show that CHE has relatively greater improvements on all metrics against BaseModels than the results with random data division, with relative increase of the averaged metric of NDCG@$k$ and ACC@$k$ by 10.06\%. And the significant test shows that, CHE+BaseModel significantly outperforms BaseModel and PW+BaseModel. 
This demonstrates our claim that the proposed CHE encourages models to rely on causal features and estimate their contributions to the prediction independently, as causal features are always useful to make disease diagnosis predictions regardless of the data distribution shifts. Moreover, the results also show that the counterfactual PW approach does not always increase the diagnosis prediction accuracy, which might be due to the inaccurate estimation of propensity scores as discussed in Section 3.2. We notice that the increase of metric values with the OOD data is more obvious for CHE+Stagenet than for other models. Based on our observation, it could be that the collaborative training of HSIC and cross-entropy loss is especially well optimized by the CHE strategy for CHE+Stagenet. We will further discuss how the CHE behaves and whether it can guide model optimization in the Visualization part.

\begin{figure}
\centering
\includegraphics[width=0.45\textwidth]{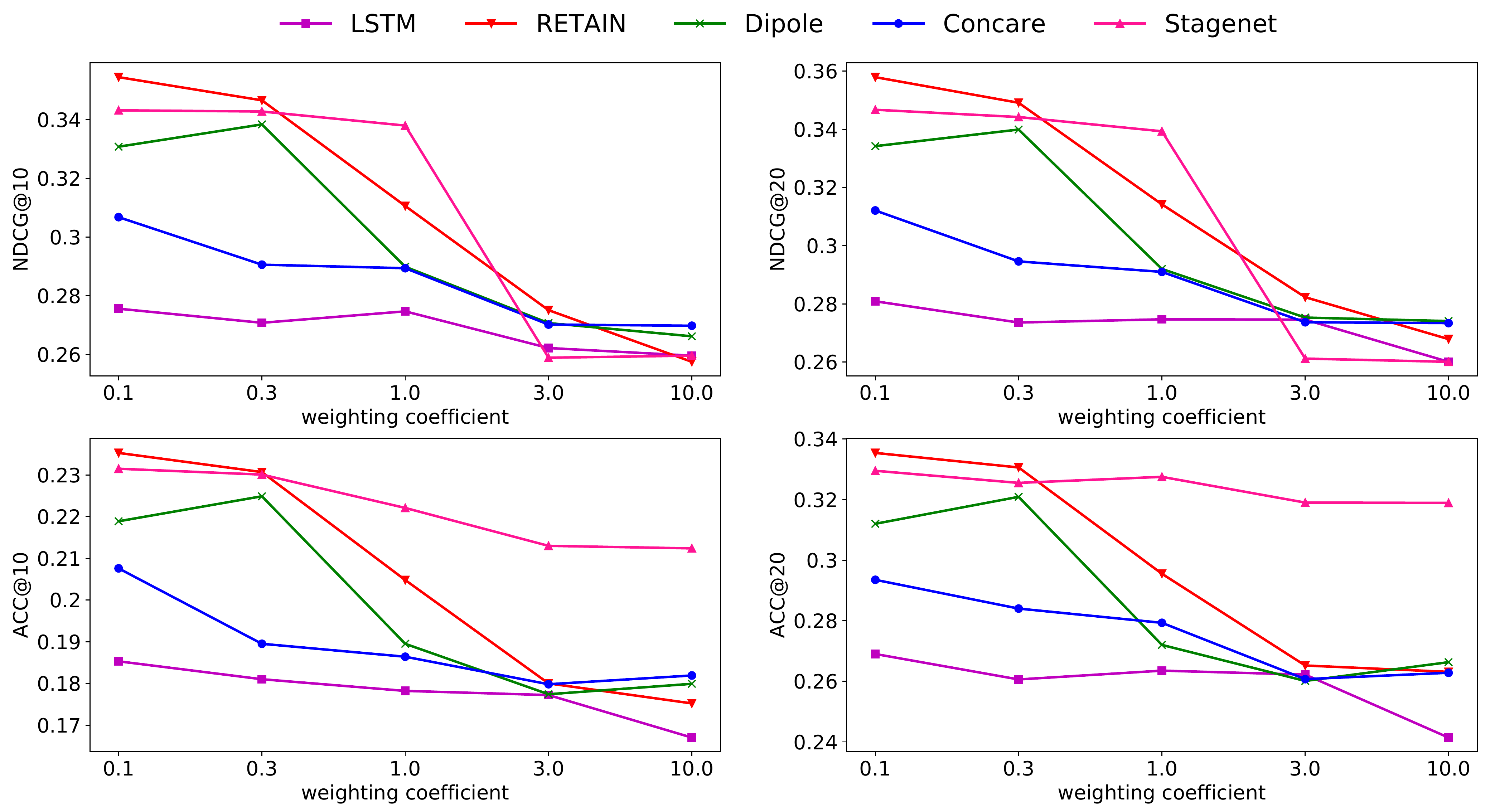}
\caption{Performances of CHE+BaseModels with different weighting coefficients on the MIMIC-III dataset with random data division.}
\label{fig:hyper1}
\end{figure}

\begin{figure}
\centering
\includegraphics[width=0.45\textwidth]{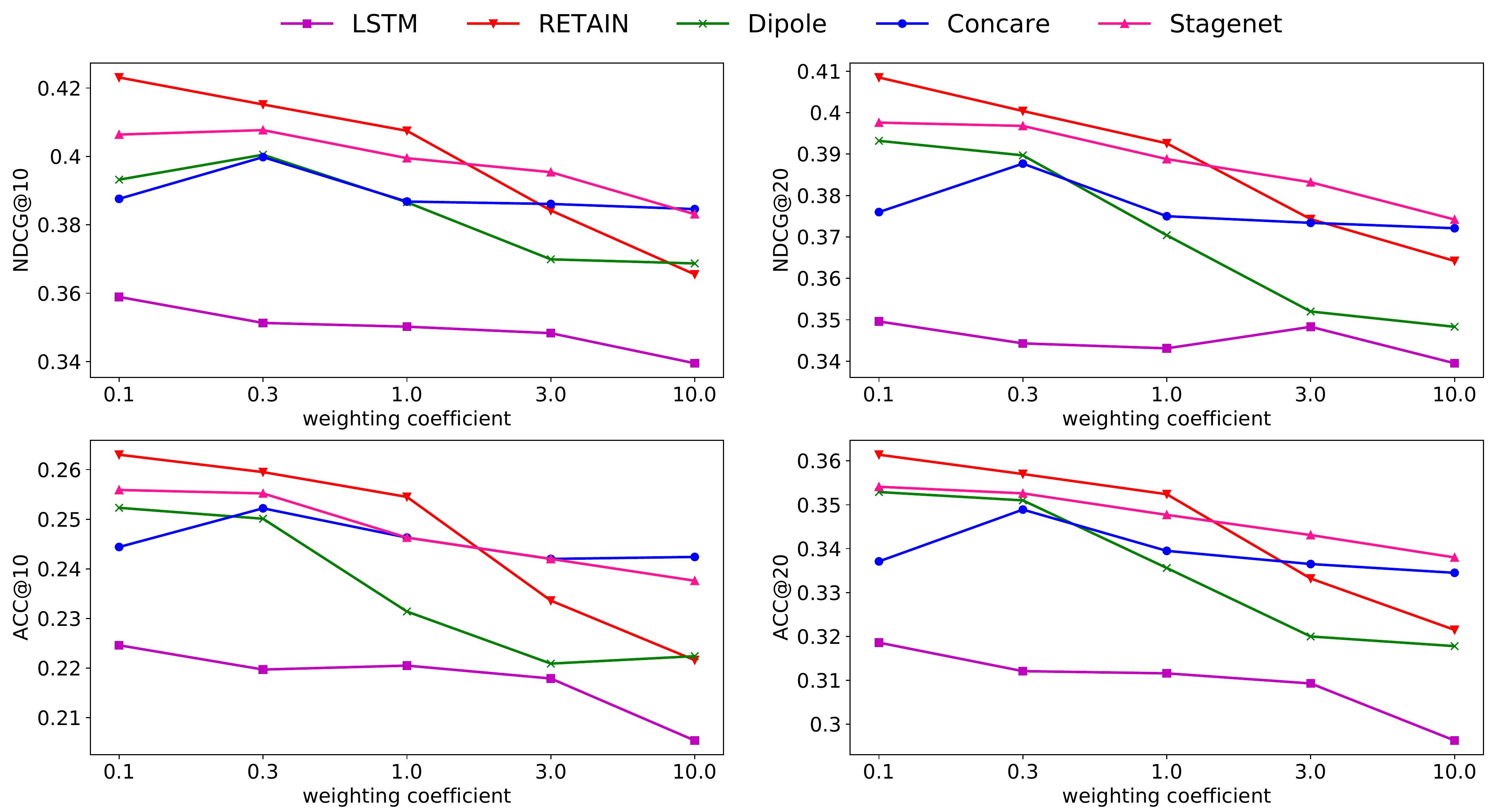}
\caption{Performances of CHE+BaseModels with different weighting coefficients on the MIMIC-IV dataset with random data division.}
\label{fig:hyper2}
\end{figure}

\subsection{Hyperparameter Study}
Because the selection of weighting coefficient $\epsilon$ is crucial for the optimization convergence, we report the model performances with different $\epsilon \in \{0.1, 0.3, 1, 3, 10\}$. Other hyperparameters are fixed. Fig. \ref{fig:hyper1} and Fig.\ref{fig:hyper2} show that smaller $\epsilon$ is better for all models on both datasets. In general, the diagnosis prediction accuracy is still sensitive to the selection of coefficient. A too-large $\epsilon$ will cause a sharp drop in prediction performance. For example, when $\epsilon \in \{3, 10\}$, the NDCG@10 and ACC@10 drop by more than 30\% for MIMIC-III dataset. $\epsilon \in \{0.1, 0.3\}$ has the optimal performance.

\subsection{Visualization}
To better understand whether the proposed Causal Healthcare Embedding can truly make the model to learn causal features, we visualize the contribution of each feature to the prediction of future diagnosis.
We know that the cause of diabetic retinopathy is diabetes. For patients with background diabetic retinopathy (ICD-9 code 36021), an ideal model should rely on related diseases such as diabetes to make prediction.
% Moreover, the contributions of each time's diagnosis and procedure should not be necessarily correlated.
In Table \ref{tab:inter}, we show the EHR of a patient in the MIMIC-III dataset and the feature interpretations, contributions of the features to the prediction, in CHE+Dipole and Dipole.
Specifically, we apply gradient backpropagation for calculating feature interpretations \cite{selvaraju2017grad,smilkov2017smoothgrad,li2015visualizing,liu2021mining}.
The contributions of diagnosis $d^i_{j'}$ and procedure $p^i_{j'}$ for predicting $d^i_{j+1}$ (${j'} \le j$) can be calculated as $\frac{{\partial \hat d_{j + 1}^i}}{{\partial d_{j'}^i}} = \frac{{\partial \hat d_{j + 1}^i}}{{\partial {\bf{E}}_d^{i,j'}}}\mathop {\left( {{\bf{E}}_d^{i,j'}} \right)}\nolimits^T$ and $\frac{{\partial \hat d_{j + 1}^i}}{{\partial p_{j'}^i}} = \frac{{\partial \hat d_{j + 1}^i}}{{\partial {\bf{E}}_p^{i,j'}}}\mathop {\left( {{\bf{E}}_p^{i,j'}} \right)}\nolimits^T$ respectively, where $\hat d_{j + 1}^i = Prd(Enc(D_j^i),Enc(P_j^i))$ is the prediction as in the loss function in Eq. (\ref{eq:loss_pred}).

\begin{table}
  \caption{Feature contributions of a patient from MIMIC-III.}
  \label{tab:inter}
  \begin{tabular}{crclrcl}
    \toprule
   \multirow{2}{*}{Feature} & \multicolumn{3}{c}{CHE+Dipole} & \multicolumn{3}{c}{Dipole}\\
    & $1^{st}$          & $2^{nd}$           & $3^{rd}$ 
    & $1^{st}$          & $2^{nd}$          & $3^{rd}$ \\
    \midrule
    Diagnosis & 0.6841 & 1.710 & 1.201 & 0.8997 & 0.1366 & 1.714\\
    Procedure & 0.3413 & 0.1777 & 0.358 & 0.5469 & 0.1733 & 1.248\\
  \bottomrule
\end{tabular}
\end{table}

\begin{figure}
\centering
\includegraphics[width=0.4\textwidth]{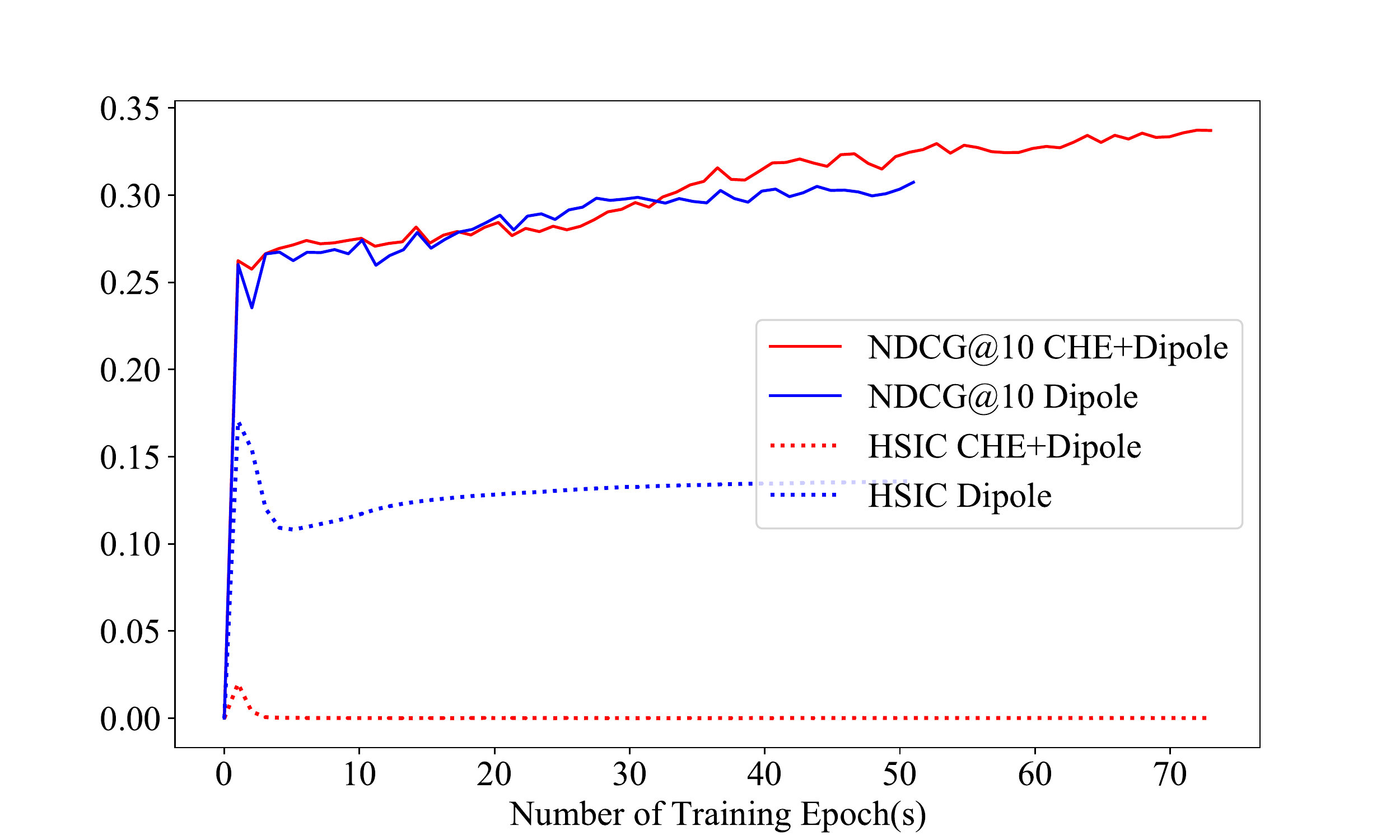}
\caption{The curves of HSIC and NDCG@10 on the test set when training CHE+Dipole and Dipole on the MIMIC-III dataset.}
\label{fig:loss}
\end{figure}

In this example, the diagnosis sequence is
\{4280, 5856\},
\{99592, 4280, 25060, 3572, V5861, V1251, 99662, 40391, 03811, 25050, 36201, 5856\},
\{03811, 5856, 99681, 42832\}.
%\{41401, 4280, 25050, 36201, 99591, 5856, 25060, 3572\}. 
The future diagnosis to be predicted is background diabetic retinopathy (36021).
The first visit contains two diseases that appear frequently among people, i.e. congestive heart failure (4280) and end stage renal disease (5856). The second visit contains some highly related features, such as diabetic retinopathy (25050), diabetes with neurological manifestations (25060), and polyneuropathy in diabetes (3572). In the third visit, the complications of transplanted kidney (99681) might be related.
Compared with Dipole, CHE+Dipole pays more attention to causal features, i.e., the second visit with many highly related diagnoses. Moreover, the contributions of diagnosis and procedure are less correlated.

We visualize the curves of HSIC and NDCG@10 on the test set in Figure \ref{fig:loss}, where the test samples are not weighted. Early stopping is adopted if the loss function does not decrease for 20 epochs. We find that CHE+Dipole has a much lower HSIC in inference and more training steps before early stopping with a higher NDCG@10 in the end. This aligns well with our observation that the CHE+Dipole prediction is more accurate, since the optimization path is guided by stable causality.

\section{Conclusion}

In this work, we focus on the problem of unstable representation learning in deep learning-based diagnosis prediction models on EHR data.
This is caused by the strong correlation between diagnoses and procedures in EHR, and it is hard for deep learning-based models to learn their causal relationships to future diagnosis.
Accordingly, we propose a CHE method to learn causal representations for diagnosis prediction models, via removing dependencies between diagnoses and procedures by weighting technique.
To be noted, our proposed CHE method can be used as a plug-and-play module.
We demonstrate by extensive experiments on the sequential diagnosis and procedure features as examples that CHE can significantly improve the performances of diagnosis prediction models.
The visualizations demonstrate that CHE is more causal than baselines and the optimization path is guided by causality, which marks a promising direction for healthcare. In future works, we will further explore the removals of spurious relationships in multimodal healthcare data for more real-world problems. 

\section{Acknowledgement}
This work is partially sponsored by CAAI-Huawei MindSpore Open Fund.

\bibliographystyle{IEEEtran}
\bibliography{ref}

\begin{thebibliography}{10}
\providecommand{\url}[1]{#1}
\csname url@samestyle\endcsname
\providecommand{\newblock}{\relax}
\providecommand{\bibinfo}[2]{#2}
\providecommand{\BIBentrySTDinterwordspacing}{\spaceskip=0pt\relax}
\providecommand{\BIBentryALTinterwordstretchfactor}{4}
\providecommand{\BIBentryALTinterwordspacing}{\spaceskip=\fontdimen2\font plus
\BIBentryALTinterwordstretchfactor\fontdimen3\font minus
  \fontdimen4\font\relax}
\providecommand{\BIBforeignlanguage}[2]{{%
\expandafter\ifx\csname l@#1\endcsname\relax
\typeout{** WARNING: IEEEtran.bst: No hyphenation pattern has been}%
\typeout{** loaded for the language `#1'. Using the pattern for}%
\typeout{** the default language instead.}%
\else
\language=\csname l@#1\endcsname
\fi
#2}}
\providecommand{\BIBdecl}{\relax}
\BIBdecl

\bibitem{charles2013adoption}
D.~Charles, M.~Gabriel, and M.~F. Furukawa, ``Adoption of electronic health
  record systems among us non-federal acute care hospitals: 2008-2014,''
  \emph{ONC data brief}, vol.~9, pp. 1--9, 2013.

\bibitem{cheng2016risk}
Y.~Cheng, F.~Wang, P.~Zhang, and J.~Hu, ``Risk prediction with electronic
  health records: A deep learning approach,'' in \emph{SDM}, 2016, pp.
  432--440.

\bibitem{ma2017dipole}
F.~Ma, R.~Chitta, J.~Zhou, Q.~You, T.~Sun, and J.~Gao, ``Dipole: Diagnosis
  prediction in healthcare via attention-based bidirectional recurrent neural
  networks,'' in \emph{KDD}, 2017, pp. 1903--1911.

\bibitem{ma2018health}
T.~Ma, C.~Xiao, and F.~Wang, ``Health-atm: A deep architecture for multifaceted
  patient health record representation and risk prediction,'' in \emph{SDM},
  2018, pp. 261--269.

\bibitem{zhang2019metapred}
X.~S. Zhang, F.~Tang, H.~H. Dodge, J.~Zhou, and F.~Wang, ``Metapred:
  Meta-learning for clinical risk prediction with limited patient electronic
  health records,'' in \emph{KDD}, 2019, pp. 2487--2495.

\bibitem{gao2020stagenet}
J.~Gao, C.~Xiao, Y.~Wang, W.~Tang, L.~M. Glass, and J.~Sun, ``Stagenet:
  Stage-aware neural networks for health risk prediction,'' in \emph{The Web
  Conference}, 2020, pp. 530--540.

\bibitem{choi2016retain}
E.~Choi, M.~T. Bahadori, J.~A. Kulas, A.~Schuetz, W.~F. Stewart, and J.~Sun,
  ``Retain: an interpretable predictive model for healthcare using reverse time
  attention mechanism,'' in \emph{NeurIPS}, 2016, pp. 3512--3520.

\bibitem{choi2017gram}
E.~Choi, M.~T. Bahadori, L.~Song, W.~F. Stewart, and J.~Sun, ``Gram:
  graph-based attention model for healthcare representation learning,'' in
  \emph{KDD}, 2017, pp. 787--795.

\bibitem{gao2019camp}
J.~Gao, X.~Wang, Y.~Wang, Z.~Yang, J.~Gao, J.~Wang, W.~Tang, and X.~Xie,
  ``Camp: Co-attention memory networks for diagnosis prediction in
  healthcare,'' in \emph{ICDM}, 2019, pp. 1036--1041.

\bibitem{ma2020concare}
L.~Ma, C.~Zhang, Y.~Wang, W.~Ruan, J.~Wang, W.~Tang, X.~Ma, X.~Gao, and J.~Gao,
  ``Concare: Personalized clinical feature embedding via capturing the
  healthcare context,'' in \emph{AAAI}, 2020, pp. 833--840.

\bibitem{huang2006correcting}
J.~Huang, A.~Gretton, K.~Borgwardt, B.~Sch{\"o}lkopf, and A.~Smola,
  ``Correcting sample selection bias by unlabeled data,'' \emph{NeurIPS}, pp.
  601--608, 2006.

\bibitem{brookhart2010confounding}
M.~A. Brookhart, T.~St{\"u}rmer, R.~J. Glynn, J.~Rassen, and S.~Schneeweiss,
  ``Confounding control in healthcare database research: challenges and
  potential approaches,'' \emph{Medical Care}, vol.~48, no. 6 0, p. S114, 2010.

\bibitem{hendrycks2018benchmarking}
D.~Hendrycks and T.~Dietterich, ``Benchmarking neural network robustness to
  common corruptions and perturbations,'' in \emph{ICLR}, 2018.

\bibitem{bengio2019meta}
Y.~Bengio, T.~Deleu, N.~Rahaman, N.~R. Ke, S.~Lachapelle, O.~Bilaniuk,
  A.~Goyal, and C.~Pal, ``A meta-transfer objective for learning to disentangle
  causal mechanisms,'' in \emph{ICLR}, 2019.

\bibitem{pearl2009causal}
J.~Pearl, ``Causal inference in statistics: An overview,'' \emph{Statistics
  Surveys}, vol.~3, pp. 96--146, 2009.

\bibitem{muandet2013domain}
K.~Muandet, D.~Balduzzi, and B.~Sch{\"o}lkopf, ``Domain generalization via
  invariant feature representation,'' in \emph{ICML}, 2013, pp. 10--18.

\bibitem{rojas2018invariant}
M.~Rojas-Carulla, B.~Sch{\"o}lkopf, R.~Turner, and J.~Peters, ``Invariant
  models for causal transfer learning,'' \emph{The Journal of Machine Learning
  Research}, vol.~19, no.~1, pp. 1309--1342, 2018.

\bibitem{peters2016causal}
J.~Peters, P.~B{\"u}hlmann, and N.~Meinshausen, ``Causal inference by using
  invariant prediction: identification and confidence intervals,''
  \emph{Journal of the Royal Statistical Society. Series B (Statistical
  Methodology)}, pp. 947--1012, 2016.

\bibitem{kuang2018stable}
K.~Kuang, P.~Cui, S.~Athey, R.~Xiong, and B.~Li, ``Stable prediction across
  unknown environments,'' in \emph{KDD}, 2018, pp. 1617--1626.

\bibitem{kuang2020stable}
K.~Kuang, R.~Xiong, P.~Cui, S.~Athey, and B.~Li, ``Stable prediction with model
  misspecification and agnostic distribution shift,'' in \emph{AAAI}, 2020, pp.
  4485--4492.

\bibitem{gretton2007kernel}
A.~Gretton, K.~Fukumizu, C.~H. Teo, L.~Song, B.~Sch{\"o}lkopf, A.~J. Smola
  \emph{et~al.}, ``A kernel statistical test of independence.'' in
  \emph{NeurIPS}, 2007, pp. 585--592.

\bibitem{greenfeld2020robust}
D.~Greenfeld and U.~Shalit, ``Robust learning with the hilbert-schmidt
  independence criterion,'' in \emph{ICML}, 2020, pp. 3759--3768.

\bibitem{bahng2020learning}
H.~Bahng, S.~Chun, S.~Yun, J.~Choo, and S.~J. Oh, ``Learning de-biased
  representations with biased representations,'' in \emph{ICML}, 2020, pp.
  528--539.

\bibitem{zou2020counterfactual}
H.~Zou, P.~Cui, B.~Li, Z.~Shen, J.~Ma, H.~Yang, and Y.~He, ``Counterfactual
  prediction for bundle treatment,'' \emph{NeurIPS}, 2020.

\bibitem{arbour2021permutation}
D.~Arbour, D.~Dimmery, and A.~Sondhi, ``Permutation weighting,'' in
  \emph{ICML}, 2021, pp. 331--341.

\bibitem{che2017boosting}
Z.~Che, Y.~Cheng, S.~Zhai, Z.~Sun, and Y.~Liu, ``Boosting deep learning risk
  prediction with generative adversarial networks for electronic health
  records,'' in \emph{ICDM}, 2017, pp. 787--792.

\bibitem{luo2022learning}
Y.~Luo, C.~Xu, Y.~Liu, W.~Liu, S.~Zheng, and J.~Bian, ``Learning differential
  operators for interpretable time series modeling,'' in \emph{Proceedings of
  the 28th ACM SIGKDD Conference on Knowledge Discovery and Data Mining}, 2022,
  pp. 1192--1201.

\bibitem{choi2015constructing}
E.~Choi, N.~Du, R.~Chen, L.~Song, and J.~Sun, ``Constructing disease network
  and temporal progression model via context-sensitive hawkes process,'' in
  \emph{ICDM}, 2015, pp. 721--726.

\bibitem{zhang2019attain}
Y.~Zhang, ``Attain: Attention-based time-aware lstm networks for disease
  progression modeling.'' in \emph{IJCAI}, 2019.

\bibitem{ye2020lsan}
M.~Ye, J.~Luo, C.~Xiao, and F.~Ma, ``Lsan: Modeling long-term dependencies and
  short-term correlations with hierarchical attention for risk prediction,'' in
  \emph{CIKM}, 2020, pp. 1753--1762.

\bibitem{che2015deep}
Z.~Che, D.~Kale, W.~Li, M.~T. Bahadori, and Y.~Liu, ``Deep computational
  phenotyping,'' in \emph{KDD}, 2015, pp. 507--516.

\bibitem{liu2015temporal}
C.~Liu, F.~Wang, J.~Hu, and H.~Xiong, ``Temporal phenotyping from longitudinal
  electronic health records: A graph based framework,'' in \emph{KDD}, 2015,
  pp. 705--714.

\bibitem{zeng2018natural}
Z.~Zeng, Y.~Deng, X.~Li, T.~Naumann, and Y.~Luo, ``Natural language processing
  for ehr-based computational phenotyping,'' \emph{IEEE/ACM Transactions on
  Computational Biology and Bioinformatics}, vol.~16, no.~1, pp. 139--153,
  2018.

\bibitem{ma2018kame}
F.~Ma, Q.~You, H.~Xiao, R.~Chitta, J.~Zhou, and J.~Gao, ``Kame: Knowledge-based
  attention model for diagnosis prediction in healthcare,'' in \emph{CIKM},
  2018, pp. 743--752.

\bibitem{choi2020learning}
E.~Choi, Z.~Xu, Y.~Li, M.~Dusenberry, G.~Flores, E.~Xue, and A.~Dai, ``Learning
  the graphical structure of electronic health records with graph convolutional
  transformer,'' in \emph{AAAI}, 2020, pp. 606--613.

\bibitem{hochreiter1997long}
S.~Hochreiter and J.~Schmidhuber, ``Long short-term memory,'' \emph{Neural
  Computation}, vol.~9, no.~8, pp. 1735--1780, 1997.

\bibitem{zhang2020inprem}
X.~Zhang, B.~Qian, S.~Cao, Y.~Li, H.~Chen, Y.~Zheng, and I.~Davidson, ``Inprem:
  An interpretable and trustworthy predictive model for healthcare,'' in
  \emph{KDD}, 2020, pp. 450--460.

\bibitem{luo2020hitanet}
J.~Luo, M.~Ye, C.~Xiao, and F.~Ma, ``Hitanet: Hierarchical time-aware attention
  networks for risk prediction on electronic health records,'' in \emph{KDD},
  2020, pp. 647--656.

\bibitem{peng2021sequential}
X.~Peng, G.~Long, T.~Shen, S.~Wang, and J.~Jiang, ``Sequential diagnosis
  prediction with transformer and ontological representation,'' \emph{arXiv
  preprint arXiv:2109.03069}, 2021.

\bibitem{zhang2021learning}
X.~Zhang, B.~Qian, Y.~Li, Y.~Liu, X.~Chen, C.~Guan, and C.~Li, ``Learning
  robust patient representations from multi-modal electronic health records: A
  supervised deep learning approach,'' in \emph{SDM}, 2021, pp. 585--593.

\bibitem{chen2021unite}
C.~Chen, J.~Liang, F.~Ma, L.~Glass, J.~Sun, and C.~Xiao, ``Unite:
  Uncertainty-based health risk prediction leveraging multi-sourced data,'' in
  \emph{The Web Conference}, 2021, pp. 217--226.

\bibitem{zhang2020hierarchical}
M.~Zhang, C.~R. King, M.~Avidan, and Y.~Chen, ``Hierarchical attention
  propagation for healthcare representation learning,'' in \emph{KDD}, 2020,
  pp. 249--256.

\bibitem{ye2021medpath}
M.~Ye, S.~Cui, Y.~Wang, J.~Luo, C.~Xiao, and F.~Ma, ``Medpath: Augmenting
  health risk prediction via medical knowledge paths,'' in \emph{The Web
  Conference}, 2021, pp. 1397--1409.

\bibitem{johansson2016learning}
F.~Johansson, U.~Shalit, and D.~Sontag, ``Learning representations for
  counterfactual inference,'' in \emph{ICML}, 2016, pp. 3020--3029.

\bibitem{morgan2015counterfactuals}
S.~L. Morgan and C.~Winship, \emph{Counterfactuals and Causal Inference}.\hskip
  1em plus 0.5em minus 0.4em\relax Cambridge University Press, 2015.

\bibitem{rosenbaum1983central}
P.~R. Rosenbaum and D.~B. Rubin, ``The central role of the propensity score in
  observational studies for causal effects,'' \emph{Biometrika}, vol.~70,
  no.~1, pp. 41--55, 1983.

\bibitem{austin2011introduction}
P.~C. Austin, ``An introduction to propensity score methods for reducing the
  effects of confounding in observational studies,'' \emph{Multivariate
  Behavioral Research}, vol.~46, no.~3, pp. 399--424, 2011.

\bibitem{hassanpour2019learning}
N.~Hassanpour and R.~Greiner, ``Learning disentangled representations for
  counterfactual regression,'' in \emph{ICLR}, 2019.

\bibitem{lopez2017estimation}
M.~J. Lopez and R.~Gutman, ``Estimation of causal effects with multiple
  treatments: a review and new ideas,'' \emph{Statistical Science}, pp.
  432--454, 2017.

\bibitem{hassanpour2019counterfactual}
N.~Hassanpour and R.~Greiner, ``Counterfactual regression with importance
  sampling weights.'' in \emph{IJCAI}, 2019, pp. 5880--5887.

\bibitem{yoon2018ganite}
J.~Yoon, J.~Jordon, and M.~Van Der~Schaar, ``Ganite: Estimation of
  individualized treatment effects using generative adversarial nets,'' in
  \emph{ICLR}, 2018.

\bibitem{shen2020stable0}
Z.~Shen, P.~Cui, J.~Liu, T.~Zhang, B.~Li, and Z.~Chen, ``Stable learning via
  differentiated variable decorrelation,'' in \emph{KDD}, 2020, pp. 2185--2193.

\bibitem{shen2020stable}
Z.~Shen, P.~Cui, T.~Zhang, and K.~Kunag, ``Stable learning via sample
  reweighting,'' in \emph{AAAI}, 2020.

\bibitem{zhang2021deep}
X.~Zhang, P.~Cui, R.~Xu, L.~Zhou, Y.~He, and Z.~Shen, ``Deep stable learning
  for out-of-distribution generalization,'' in \emph{CVPR}, 2021, pp.
  5372--5382.

\bibitem{fan2021generalizing}
S.~Fan, X.~Wang, C.~Shi, P.~Cui, and B.~Wang, ``Generalizing graph neural
  networks on out-of-distribution graphs,'' \emph{arXiv preprint
  arXiv:2111.10657}, 2021.

\bibitem{farrar1967multicollinearity}
D.~E. Farrar and R.~R. Glauber, ``Multicollinearity in regression analysis: the
  problem revisited,'' \emph{The Review of Economic and Statistics}, pp.
  92--107, 1967.

\bibitem{sriperumbudur2009kernel}
B.~K. Sriperumbudur, K.~Fukumizu, A.~Gretton, G.~R. Lanckriet, and
  B.~Sch{\"o}lkopf, ``Kernel choice and classifiability for rkhs embeddings of
  probability distributions.'' in \emph{NIPS}, vol.~22, 2009, pp. 1750--1758.

\bibitem{johnson2016mimic}
A.~E. Johnson, T.~J. Pollard, L.~Shen, H.~L. Li-Wei, M.~Feng, M.~Ghassemi,
  B.~Moody, P.~Szolovits, L.~A. Celi, and R.~G. Mark, ``Mimic-iii, a freely
  accessible critical care database,'' \emph{Scientific Data}, vol.~3, no.~1,
  pp. 1--9, 2016.

\bibitem{selvaraju2017grad}
R.~R. Selvaraju, M.~Cogswell, A.~Das, R.~Vedantam, D.~Parikh, and D.~Batra,
  ``Grad-cam: Visual explanations from deep networks via gradient-based
  localization,'' in \emph{ICCV}, 2017, pp. 618--626.

\bibitem{smilkov2017smoothgrad}
D.~Smilkov, N.~Thorat, B.~Kim, F.~Viegas, and M.~Wattenberg, ``Smoothgrad:
  removing noise by adding noise,'' in \emph{ICML}, 2017.

\bibitem{li2015visualizing}
J.~Li, X.~Chen, E.~Hovy, and D.~Jurafsky, ``Visualizing and understanding
  neural models in nlp,'' \emph{arXiv preprint arXiv:1506.01066}, 2015.

\bibitem{liu2021mining}
Q.~Liu, Z.~Liu, H.~Zhang, Y.~Chen, and J.~Zhu, ``Mining cross features for
  financial credit risk assessment,'' in \emph{CIKM}, 2021.

\end{thebibliography}
\vspace{12pt}
\color{red}

\end{document}